\documentclass[runningheads,a4paper]{llncs}
\usepackage{multirow}

\def\RR{{\rm I\hspace{-0.50ex}R}}

\def\esp{\hspace*{.1in}}
\def\NF{{Displacement}}
\def\LS{{Novelty}}

\usepackage{amssymb}
\usepackage{graphicx}

\usepackage{url}

\begin{document}

\mainmatter  

\title{Open-Ended Evolutionary Robotics:\\ an Information Theoretic Approach}


\titlerunning{Information Theory for Open-Ended Evolutionary Robotics}

%
%
\author{Pierre Delarboulas\footnote{funded by EU FP7, FET IP SYMBRION No. 216342, http://symbrion.eu/} \and Marc Schoenauer \and Mich\`ele Sebag}
\authorrunning{Delarboulas, Schoenauer, and Sebag}

\institute{Project-team TAO $-$ LRI, UMR CNRS 8623 \& INRIA Saclay\\
Universit\'e Paris-Sud, F-91405 Orsay}
\date{\{delarboulas,marc,sebag\}@lri.fr}

\toctitle{~}
\tocauthor{~}
\maketitle

\begin{abstract}
This paper is concerned with designing self-driven fitness functions for 
Embedded Evolutionary Robotics. The proposed approach considers the entropy of the 
sensori-motor stream generated by the robot controller. This entropy is computed using 
unsupervised learning; its maximization, achieved by an on-board evolutionary algorithm, 
implements a ``curiosity instinct'', favouring controllers visiting many diverse 
sensori-motor states (sms).
Further, the set of sms discovered by an individual can be transmitted to its offspring, making 
a cultural evolution mode possible. Cumulative entropy (computed from ancestors and current individual 
visits to the sms) defines another self-driven fitness; its optimization implements a ``discovery
instinct'', as it favours controllers visiting new or rare sensori-motor states. 
Empirical results on the benchmark problems proposed by Lehman and Stanley (2008) comparatively
demonstrate the merits of the approach.
\end{abstract}

\section{Introduction}
Evolutionary Robotics (ER) aims at designing
robust autonomous robots, and in particular robust robot controllers \cite{Nolfi:Floreano:Book00}. 
The success of ER critically depends on the optimization objective (or fitness 
function) \cite{Nelson}. For the sake of computational and experimental convenience, 
the ER literature mostly considers 
simulation-based approaches, 
where the 
controller fitness is computed by simulating the robot behaviour. The price to 
pay for this convenience is that the controller behaviour suffers from the so-called 
reality gap, i.e. its performance might dramatically decrease when ported on-board \cite{LipsonLegged}. 
While Embedded Evolutionary Robotics \cite{Bredeche-Haasdijk-Eiben-2009} sidesteps the 
reality gap as evolutionary computation is achieved on-board, the challenge is 
to design some fitness either based on environmental cues (e.g. about the 
target location), or not requiring any ground truth at all, i.e.,
self-driven fitness. 

Developmental Robotics \cite{PfeiferEtAl07} also
aims at principled ways of building intelligent agents.
The vision, strongly inspired from Brooks' \cite{Brooks1991Intelligence-Wi},
states that (i) the world is its best model and the robot representations
must be grounded in its physical perceptions; (ii) robots must be
``autonomous, self-sufficient, embodied, and situated''
(Complete Agent principle). 
The search for self-driven objectives, leading the robot to 
explore its world and gradually learn new skills, thus is at the core of Developmental 
Robotics \cite{Oudeyer2}.

The present work, at the crossroad of Developmental and Embedded ER, 
focuses on self-driven fitness functions. An original approach,  amenable to on-board evolutionary optimization and rooted in 
Information Theory, is presented as an Intrinsic Motivation Systems \cite{Oudeyer2}.
This approach defines a ``curiosity'' instinct, which enforces the exploration by the robot of 
its environment. Both approaches exploit the robotic log recording for each time
step the sensor and motor values, or Sensori-Motor Stream (SMS).
The difference is twofold. On the one hand, the presented approach relies on 
(computationally frugal) unsupervised learning \cite{Duda2001} whereas \cite{Oudeyer2} uses
supervised ML to build a forward model. On the other hand, it defines a representation
of the robot world: a set of sensori-motor states (sms), built from the SMS, supports the 
computation of the SMS entropy; the higher the entropy, the richer and the more diversified
the world seen by the current controller.   
Maximizing the SMS entropy thus defines an efficient, self-driven, ``curiosity instinct'' enforcing the exploration of 
the world. 

The second contribution of the paper is another self-driven fitness function dubbed ``discovery instinct''.
It exploits the fact that sensori-motor states (sms) can be transmitted 
from parents to offspring, thus enabling some cultural evolution mode \cite{curran2007,GuzsXX}. Formally,
the discovery fitness computes the cumulative entropy defined from the visits of all robots 
to all sms, until the current individual; it thus rewards individuals that
discover sensori-motor states {\em which have not yet been visited} (or rarely visited) 
by its ancestors.
Discovery fitness might be thought of as a fitness sharing mechanism, except for the fact that
it rewards individuals which differ from their ancestors, as opposed to, their peers. 

The merits of the two self-driven fitness functions, implemented within an embedded (1+1)-Evolution Strategy 
\cite{Bredeche-Haasdijk-Eiben-2009}, are comparatively assessed on the benchmark problems 
defined by Lehman and Stanley \cite{Lehman}, in terms of their patrolling activity (visiting the various places of
the arena, and visiting the chambers farthest apart from the starting point). The main limitation
of the approach is to require a stimulating interaction between the 
environment and the robot sensors, conducive to diversified sensory experiences:
if located in the middle of nowhere, or endowed with too poor sensors, 
the robot can only experience a few sensori-motor states and entropy provides no incentive 
for exploration.

The paper is organized as follows. Related work is briefly reviewed and discussed in section 
\ref{SoA}. An overview of the curiosity and discovery fitnesses is presented in section 
\ref{instinct} and section \ref{expe} reports on the experimental results. The paper concludes with a 
discussion and some perspectives for further research.

\section{Related work}\label{SoA}
\begin{figure}[htbp]
\centerline{\begin{tabular}{cc}
\includegraphics[width=0.4\textwidth]{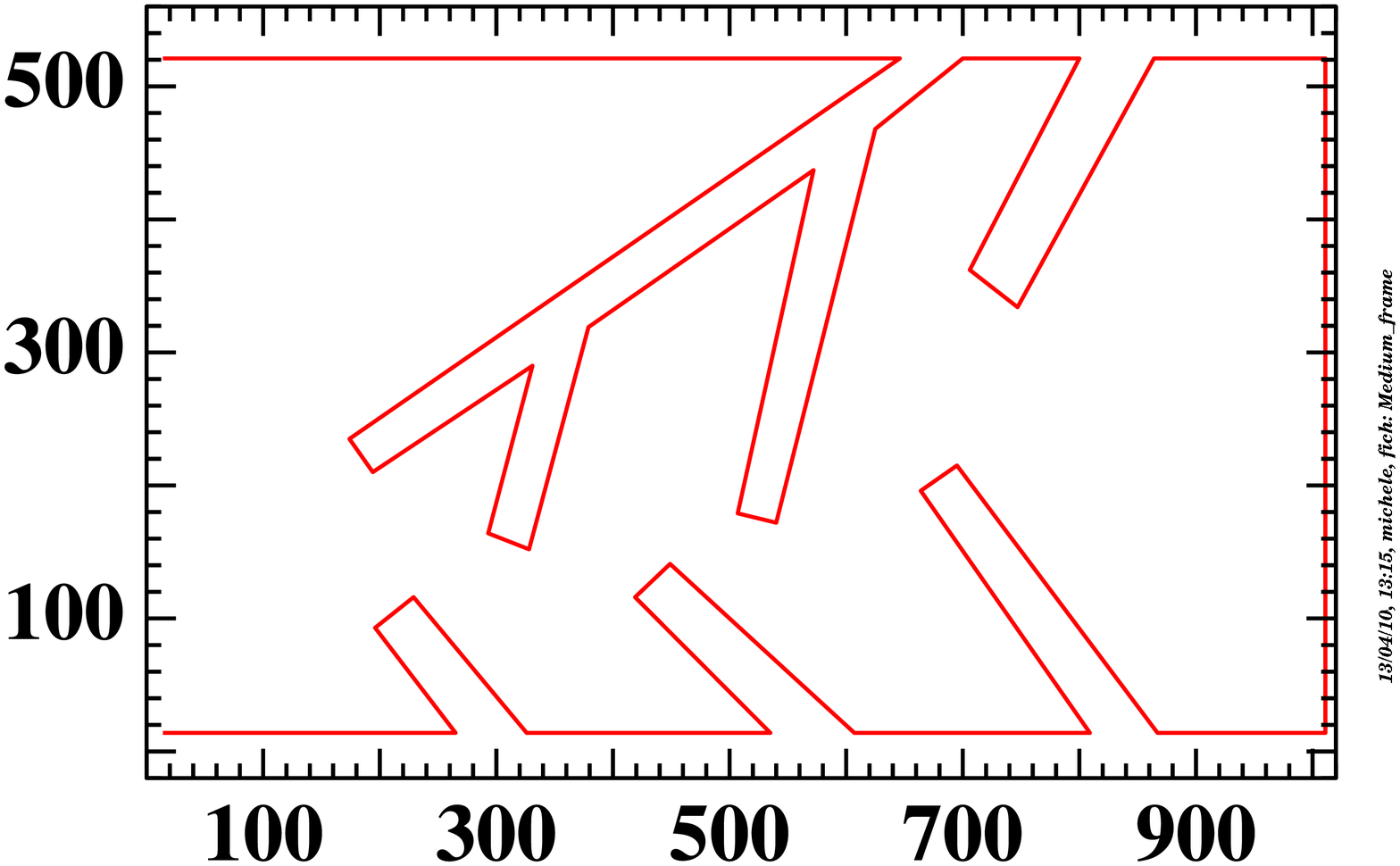} &
\includegraphics[width=0.4\textwidth]{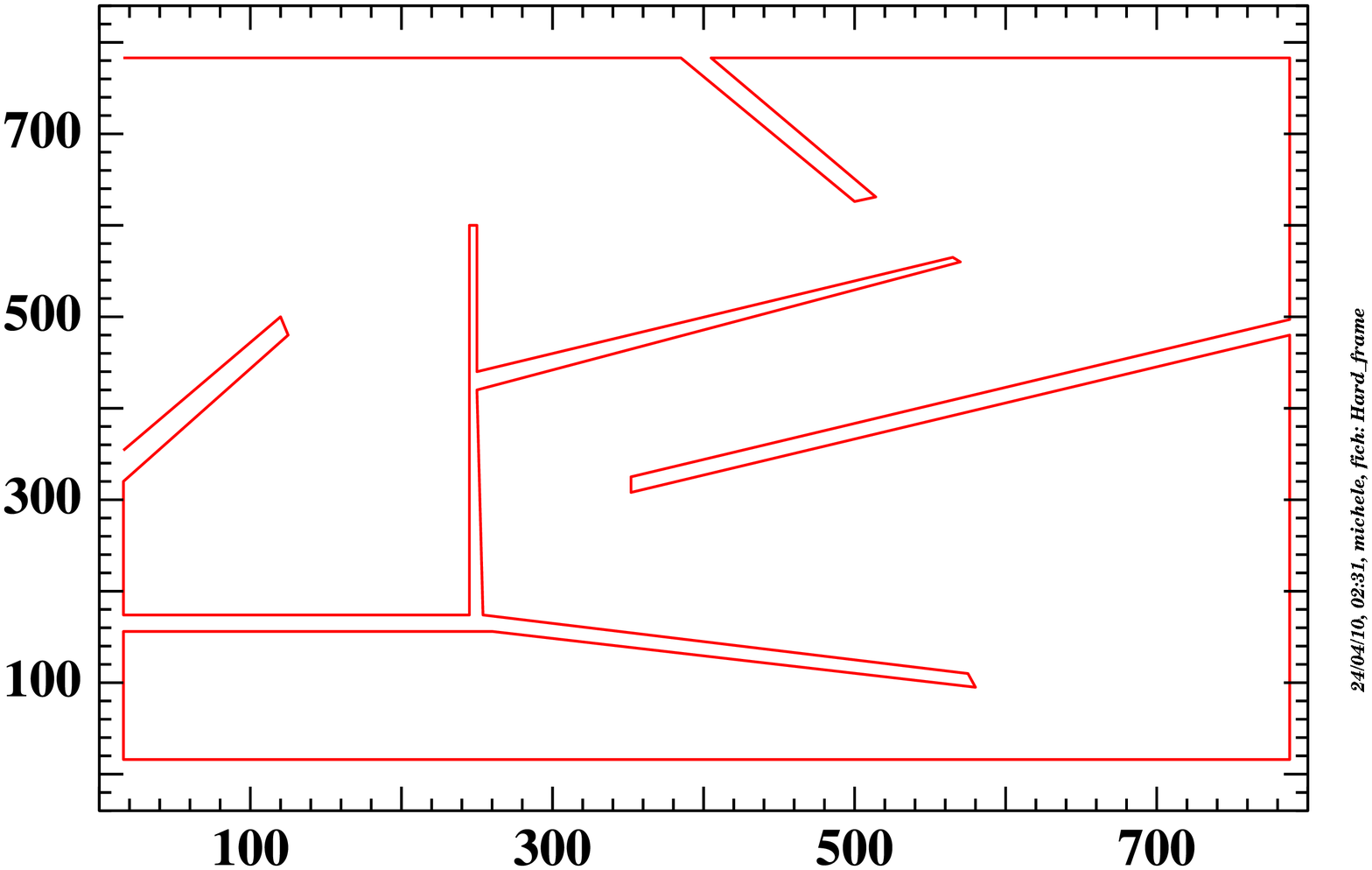}\\
\end{tabular}}
\caption{From (Lehman \& Stanley, 08): medium (left) and hard (right) arenas. 
Starting in the upper left corner, the goal is to reach the farthest apart region.
}
\label{fig:leh}
\end{figure}

The state of the art in ER and fitness design can be structured in different ways \cite{Watson02,Nelson}, depending 
on the designer's criteria. The perspective presented in \cite{Nelson}
focuses on the amount of human effort and prior knowledge required
 to overcome bootstrap problems. Such problems can be illustrated from  
the hard and medium arenas due to \cite{Lehman} 
(Fig. \ref{fig:leh}), where the goal is to reach the chamber farthest apart from the 
starting point 
without any ground truth being available; note that providing the robot with its 
bird eye's distance to the target location will get it trapped in many local optima. 

Among the various heuristics investigated to address the bootstrap problem is the 
staged fitness approach pioneered by \cite{Harvey} (e.g. learning to walk before learning to run). 
Among the early and still widely used\footnote{Other possibilities, outside the 
scope of this paper, are to inject competent individuals in the initial population, or to use
co-evolution.}
approaches to bootstrap avoidance are  
fitness-sharing and diversity enforcing \cite{Lehman,Gomez}.

Diversity enforcing heuristics in ER mostly rely on the genotypic \cite{NEAT} or phenotypic 
\cite{Gomez,Lehman} distances between the robot controllers. Some genotypic distances 
have been defined on structured controller spaces such as neural nets \cite{NEAT}. 
Phenotypic distances usually rely on prior knowledge. For instance 
Lehman and Stanley associate to a controller the 
end point of the robot trajectory; the novelty of a controller w.r.t. the population
is then assessed from the average distance of its end point to that of its $k$-nearest neighbours \cite{Lehman}. 
A general phenotypic distance between robot trajectories has been proposed by Gomez, relying on the compression-based distance inspired from 
Kolmogorov complexity \cite{Gomez}. Setting the fitness function to the controller phenotypic diversity leads
evolution to construct a fair sample of the robot behavioural space.
In this sample, the designer eventually selects the individual best fitting
the task at hand, which reportedly gives satisfying solutions.
Another way of enforcing diversity is based on multi-objective evolutionary optimization, 
considering diversity as an additional objective besides 
the actual robotic objective \cite{Mouret09}.

Another influential line of research is Embodied Statistical Learning (ESL) 
 \cite{PfeiferEtAl07}, aimed at a principled way of addressing the reality gap issue
through perceptual and behavioural learning. While perceptual
learning is concerned with building a grounded representation of the environment and
the robot itself, behavioural learning aims at achieving the target tasks.
Regarding perceptual learning, ESL advocates the use
of the ''information self-structuring`` principle, stating that the
robot should take advantage of statistical regularities induced by its
interactions with the world, and mentions ''adaptive compression`` 
as a possible approach \cite{Lungarella-08}.

\section{Information Theory-based Robotic Instincts}\label{instinct}
This section presents two self-driven fitness functions, referred to as Curiosity 
and Discovery instincts, simultaneously enforcing the 
exploration of the environment and the behavioural diversity of the controllers. 
These functions, also rooted in the information self-structuring principle, are
based on unsupervised statistical learning. 

\subsection{Unsupervised Learning from the Sensori-Motor Stream}
Some information the robot gets for free lies in its sensor and motor values. Let the sequence of sensori-motor value vectors along time,
referred to as Sensori-Motor Stream (SMS), be denoted ${\cal X} = \{x_t ; x_t \in \RR^d, 
t=1, \ldots, T\}$, where vector $x_t$ stores the $d$ values of the sensors plus motors at time $t$, and $T$ is the 
number of time steps of the robot lifetime. The SMS ${\cal X}$ of course depends on the controller of the robot when it is gathered.
Our claim is that 
{\em interesting controllers result in a high sensori-motor diversity} (subject to 
requirements discussed in section \ref{s:efdiscu}).
After Information Theory principles, 
it thus comes naturally to assess the controller from the entropy
of the SMS. Noting $(c_i)_{i = 1, \ldots, p}$ the $p$ states (sensori-motor vectors) visited by 
the robot and $n_i$ the number of times $c_i$ has been visited, it comes:

\begin{equation}
 {\cal F}({\cal X}) = - \sum_{i=1}^p \frac{n_i}{\sum_{j=1}^p n_j} \log{ \frac{n_i}{\sum_{j=1}^p n_j} } 
 \label{eqf}
\end{equation}
Sensori-motor states should however achieve some abstraction or generalization 
relatively to the sensori-motor vectors. Otherwise, since $x_t$ is a real-valued vector 
in a possibly high dimension space, for most visited states $n_i = 1$  and most trajectories
over $T$ time steps get the same trivial fitness value $\log(T)$.
Unsupervised learning (clustering), is applied to the robotic log to form clusters 
using Euclidean distance. 
While many clustering algorithms have been designed in the literature, only the simple, 
computationally linear $k$-means and $\epsilon$-means will be considered in the 
rest of this paper (Fig. \ref{alg-k}, left); the interested reader is referred to \cite{Duda2001} for a more comprehensive 
introduction. The $k$-means algorithm is parametrized from the number $k$ of clusters
while $\epsilon$-means is parametrized from the maximal radius $\epsilon$  of each cluster.

\begin{figure}[htbp]
\begin{tabular}{cc}
\parbox{.5\linewidth}
{\footnotesize
\begin{tabular}{l}
 {\bf $k$-means Algorithm}\\
 ${\cal C} = {c_1 \ldots c_k}$ random training points \\
 repeat\\
 \esp   for t = 1\ldots T, \\
 \esp \esp $i(t) = \, argmin_{j=1\ldots k} \{ d(x_t,c_j) \}$\\
 \esp   for i = 1 \ldots k  \\
 \esp \esp $c_i = \frac{\sum_{t/ i(t) = i} x_t}{\sum_{t/ i(t) = i}1}$\\
 until $\cal C$ does not change
\end{tabular}}
&
\parbox{.45\linewidth}
{\footnotesize
\begin{tabular}{l}
 {\bf $\epsilon$-means Algorithm }\\ 
 ${\cal C} = \emptyset$\\
 for t = 1..T\\
 \esp $ i(t) = \, argmin_{c_j \in {\cal C}} \{ d(x_t,c_j) \}$\\
 \esp if ($d(x_t,c_i) > \epsilon$) 
 \esp \esp ${\cal C} \leftarrow {\cal C} \bigcup (x_t,1)$\\
 \esp   else $n_i++$\\
\end{tabular}}
\end{tabular}

\caption{$k$-means and $\epsilon$-means clustering algorithms.
}
\label{alg-k}
\end{figure}


\subsection{Curiosity and Discovery Instincts}\label{s:cul}
Let $X$ be a controller. The rest of the paper does not depend on the actual representation
of $X$ (its genotype), and only considers its phenotype ${\cal X} = \{ x_1\ldots x_T\}$,
defined as its SMS in the 
environment. Stream $\cal X$ is processed using $k$-means
or $\epsilon$-means (Fig. \ref{alg-k}), yielding the set of $p$ sensori-motor states $c_i$
together with their number $n_i$ of occurrences in the stream. 
The curiosity fitness ${\cal F}_c$ is defined as 
the entropy of the trajectory (Eq. (\ref{eqf})).
An individual gets a high curiosity fitness if it equally shares its time among the 
visited sms ($k$-means clustering, $p = k$), or it visits many sms 
($\epsilon$-means clustering).
The maximization of the curiosity fitness is achieved using a $(1+1)$-Evolution Strategy with random
restart (Section \ref{expe}).

As will be discussed in Section \ref{s:efdiscu}, the number $p$ of sensori-motor states
must be kept below a few hundreds for the sake of efficiency. This makes it feasible 
to store the corresponding set of sms on-board, and to transmit it from the parent to the offspring. 
Another self-driven fitness function dubbed ``discovery instinct'' can thus be defined. Informally,
the idea is that the offspring will be rewarded for visiting sensori-motor states {\em which have
not been visited} (or rarely visited) by its ancestors. 

Formally, the discovery fitness ${\cal F}_d$ 
is defined along the same equation as the curiosity fitness. The only difference is 
that the $\epsilon$-means algorithm uses the 
set of sms $c_i$ visited by ancestors, where $n_i$ stands for the total number of visits paid to state $c_i$ 
along the generations, to initialize $\cal C$; $\cal C$ is updated from the current trajectory, incrementing the 
counters of visited states and possibly adding new sensori-motor states discovered by the 
current individual.
Ultimately, the discovery fitness ${\cal F}_d$ is computed from the entropy of $\cal C$ (Eq. 
(\ref{eqf})).

The discovery fitness thus implements a dynamic evolution schedule: the
worth of any given behaviour depends on its novelty, pushing evolution toward the collective exploration
of the sensori-motor space. Typically, an individual controller will get a good discovery fitness iff
it discovers new sms, or if it visits sufficiently many rare sms, where novelty and rarity are 
measured from the current robot-kind experience. As already noted, discovery can thus be viewed
in terms of fitness sharing, as the worth of visiting a sensori-motor state depends 
on how many individuals visited it. The difference compared with standard fitness sharing is
that sharing usually takes place among all individuals in a same generation, whereas discovery
considers all generations up to the current one\footnote{Discovery fitness can thus also
be thought of in terms of Cultural Evolution. However, the knowledge gained in the previous 
generations only modifies the individual assessment in the proposed scheme, contrasting with modifying
the individual behaviour in standard Cultural Evolution (see e.g. \cite{curran2007,GuzsXX}).}.

\subsection{Discussion}\label{s:efdiscu}
The main limitation of the proposed approach is as follows. The entropy of the 
robot trajectory depends on the richness of both the environment and the robot sensors. 
If the robot is in the middle of an empty area, there is nothing to be curious about and nothing
to be discovered; whatever its sensors, the robot will experiment a single state:
nothing in sight\footnote{
The implicit assumption done in the paper being that the controller is deterministic.}. 
Likewise, if the robot is endowed with a single boolean touch sensor, whatever the richness of
the environment the robot can 
only experience two states: I can touch something, or I can't. The presented entropy-based 
approach thus only makes sense if the environment mediated by the robot sensors offers sufficient
stimulation.

Another critical aspect is the calibration of the clustering algorithm (parameters $k$ or $\epsilon$, 
Fig. \ref{alg-k}). At one extreme (too fine-grained) all sensori-motor vectors belong to 
different clusters; at the other extreme (too coarse), all belong to the same cluster; in 
both cases, the entropy is trivial and does not provide any indication to evolution.
Along the same lines, rich robotic sensors (e.g. a camera) could hinder the approach 
due to the curse of dimensionality, and the fact that Euclidean distance in $\RR^D$ is not
much informative for high $D$ values. A preliminary dimensionality reduction step, mapping $\RR^D$ onto $\RR^d, d << D$, would thus be required, and could be obtained by careful sensor fusion. It must
however be noted that, provided that the dimensionality reduction can be done online, its
calibration (e.g. using Principal Component Analysis or non-linear approaches) can be 
optimized off-line; the approach thus remains tractable in the context of embedded evolution.

The Curiosity fitness can possibly reward some degenerate behaviours, like dancing in front of 
a wall or in a corner; more generally, a periodic trajectory in a stimulating environment would
get a high Curiosity fitness. The Discovery fitness is less prone to degenerate behaviours
since it essentially rewards new behaviours. 

Conditionally to a stimulating environment and a reasonably calibrated clustering algorithm
(in practice, a few hundred sensori-motor states), 
the Curiosity and Discovery fitnesses display interesting properties.
First of all, they meet the on-board evolution requirements (bounded computational
and memory resources, no ground truth required). Secondly, they are robust w.r.t 
sensor and motor noises; introducing outliers in the SMS would result in creating sms that are very rarely visited, with little impact on the entropy. 
Thirdly, these fitnesses penalize inaction and favour the exploration of the sensori-motor
space (not moving leads the robot to experience a single sensori-motor state). 

\section{Experimental validation}\label{expe}
The main two questions investigated in this section are (i) whether 
the curiosity and discovery fitnesses are actually compatible with on-board evolution, overcoming
the reality gap issue; and (ii) whether they are conducive to the discovery of ``interesting''
behaviours, measured from the exploration of sufficiently complex arenas.
Other questions of interest, regarding the sensitivity of the approach w.r.t.  the clustering parameters, will not be addressed here due to space limitations: only the $\epsilon$-clustering, with $\epsilon = .2$ for Curiosity and $\epsilon = .4$ for Discovery,
will be presented below. 

The experimental setting 
is based on the home-made Roborobo 2D simulator, simulating a Cortex-M3 with eight infra-red sensors and two motors: 
Following \cite{Bredeche-Haasdijk-Eiben-2009}, the robot supports an on-board (1+1)-Evolution Strategy using the $1/5^{th}$ rule, with restart after 30 fitness evaluations with no improvement;
each run stops after 2,000 fitness evaluations. The controller space is that of multi-layer perceptrons with 8 inputs, 2 outputs, and 10 hidden neurons. The 112 weights are initially randomly drawn following a normal distribution with mean $0$ and variance $0.1$. The isotropic Gaussian mutation has an initial step-size of $0.2$. In both settings, the sensori-motor stream is clustered online 
using $\epsilon$-means, with $\cal C$ initialized to the empty set for Curiosity, and, for Discovery, to the inherited set, easily stored on the Cortex board. The entropy of $\cal C$ is computed after $T=2000$ time steps. 

The robot environment is set to one of the arenas defined in
\cite{Lehman} (Fig. \ref{fig:leh}).
The performances of Curiosity and Discovery fitnesses are compared, with same
experimental setting, to both the 
baseline fitness for displacement with obstacle avoidance originally proposed by Floreano and Mondada (and described, with references, in  \cite{Nolfi:Floreano:Book00})
(legend {\em \NF}), 
and the Novelty fitness proposed by Lehman and Stanley \cite{Lehman} (see Section \ref{SoA}, legend {\em \LS}). The performance indicators measure the patrolling
ability, that is the percentage of $p(\ell)$ of squares in the arena that have been 
visited at least $\ell$ times. Another performance indicator is whether the 
robot can explore the chambers farthest apart (avoiding obstacles) from the starting point. 
All robots start from the same point, in order to reliably assess the robustness of the algorithm w.r.t. the distance to the starting point.
All results are averaged over 11 independent runs.


\begin{table}[ht!]
\centerline{\footnotesize
  \begin{tabular}{|c|c|c|c||c|c|c|}\hline
 & \multicolumn{3}{|c||}{Medium Arena} & \multicolumn{3}{|c|}{Hard Arena} \\ \hline
 & 2 visits & 5 visits & 10 visits & 2 visits & 5 visits & 10 visits \\ \hline
 \multicolumn{7}{|c|}{100 best individuals in 2000 generations run } \\ \hline
Curiosity & $ 35.78  (9.04)$  &$ 22.01  (7.39)$  &$ 13.0  (4.47)$ & $ 50.18  (6.7)$  &$ 30.31  (5.79)$  &$ 14.79  (3.16)$  \\ \hline
Discovery & $ 21.99  (9.24)$  &$ 12.82  (5.85)$  &$ 8.12  (3.15)$ & $ 16.28  (6.27)$  &$ 10.11  (3.19)$  &$ 7.26  (1.83)$  \\ \hline
\NF & $ 25.78  (1.77)$  &$ 22.3  (1.96)$  &$ 18.12  (1.86)$  & $ 44.9  (10.51)$  &$ 28.22  (6.2)$  &$ 18.99  (3.79)$  \\ \hline
\LS & $ 53.99  (2.75)$  &$ 36.32  (2.26)$  &$ 21.03  (1.61)$ & $ 55.35  (4.66)$  &$ 35.87  (3.35)$  &$ 19.78  (2.0)$  \\ \hline
\hline
  \multicolumn{7}{|c|}{All individuals in 2000 generations run } \\ \hline
Curiosity &$ 69.67  (2.62)$  &$ 61.56  (2.87)$  &$ 54.95  (3.36)$  & $ 78.29  (5.12)$  &$ 68.32  (5.58)$  &$ 58.09  (4.47)$  \\ \hline
Discovery &$ 62.08  (3.47)$  &$ 53.0  (4.59)$  &$ 45.54  (5.81)$ & $ 65.23  (4.98)$  &$ 52.4  (5.59)$  &$ 42.27  (4.97)$  \\ \hline
\NF &$ 72.36  (2.23)$  &$ 61.06  (3.59)$  &$ 50.92  (1.8)$  & $ 78.48  (3.93)$  &$ 64.98  (5.24)$  &$ 52.63  (3.68)$  \\ \hline
\LS &$ 64.82  (1.81)$  &$ 55.68  (1.76)$  &$ 49.23  (1.67)$  & $ 66.37  (5.08)$  &$ 55.31  (4.53)$  &$ 47.24  (3.75)$  \\ \hline
\end{tabular}   }
\caption{Patrolling performances for 2000 time steps: average (std. dev.) over 11 runs.}
\label{tab:patrol}
\end{table}

\begin{figure}[b!]
\centerline{\begin{tabular}{cccc}
\includegraphics[width=1.1in]{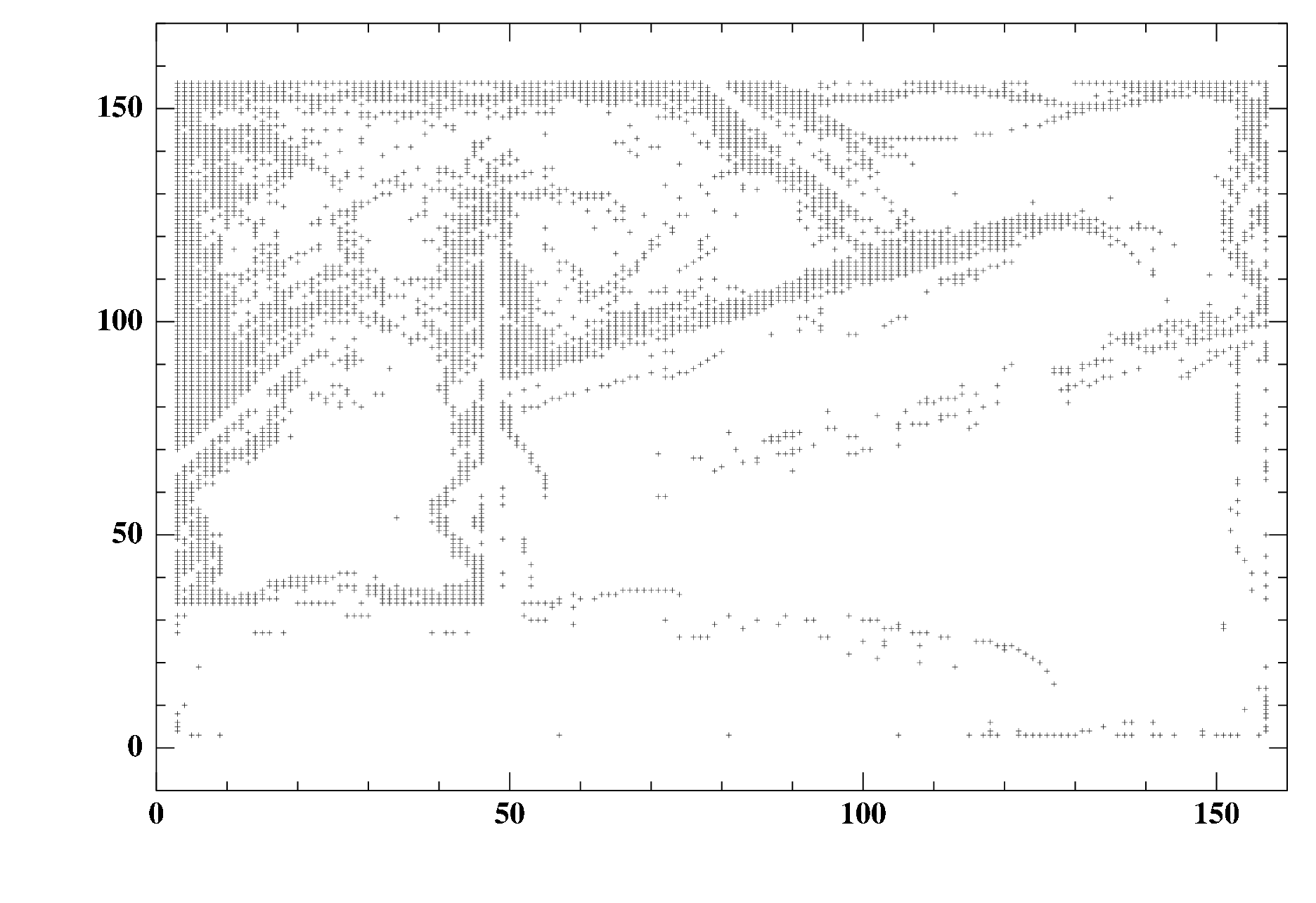} &
\includegraphics[width=1.1in]{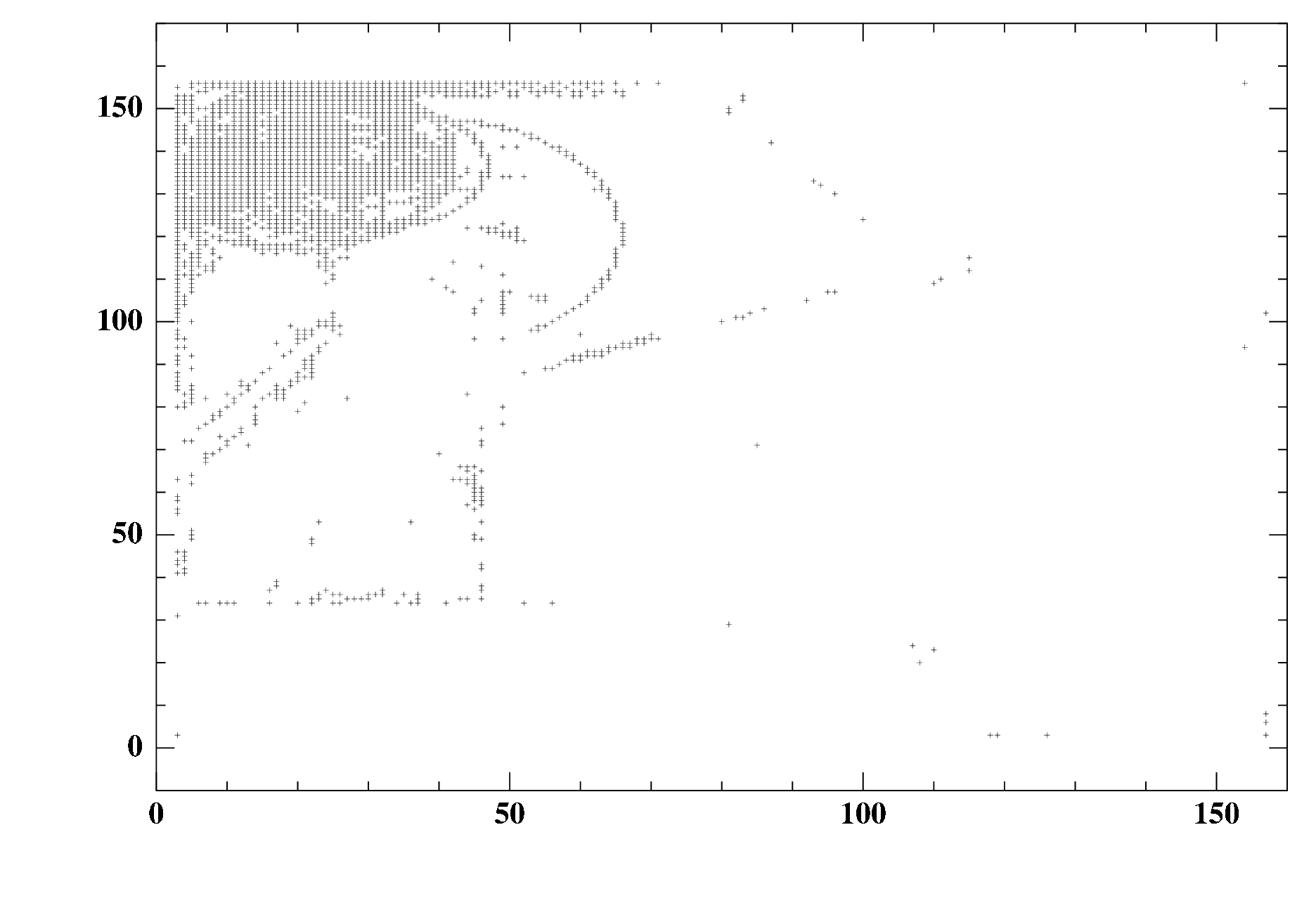} &
\includegraphics[width=1.1in]{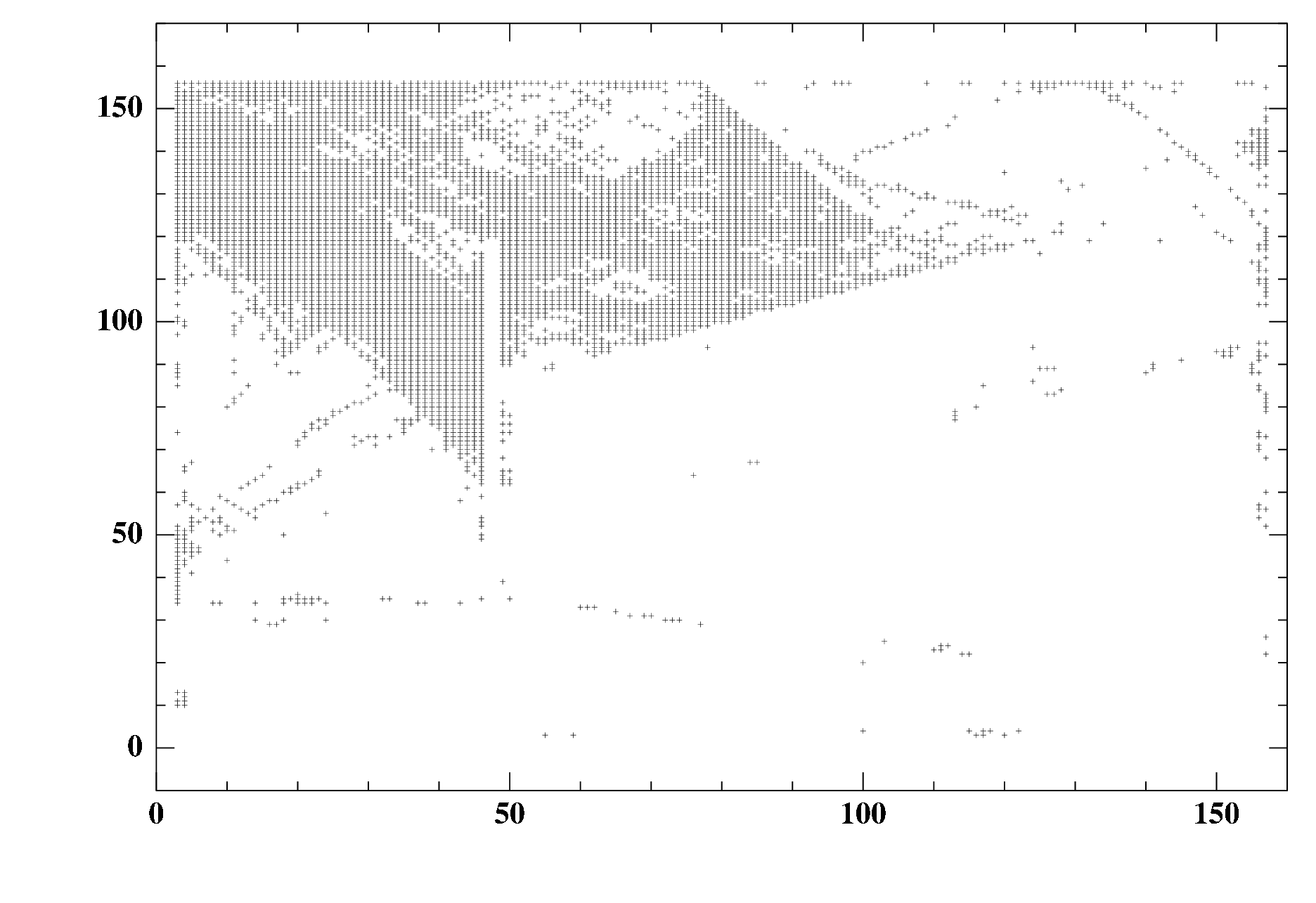} &
\includegraphics[width=1.1in]{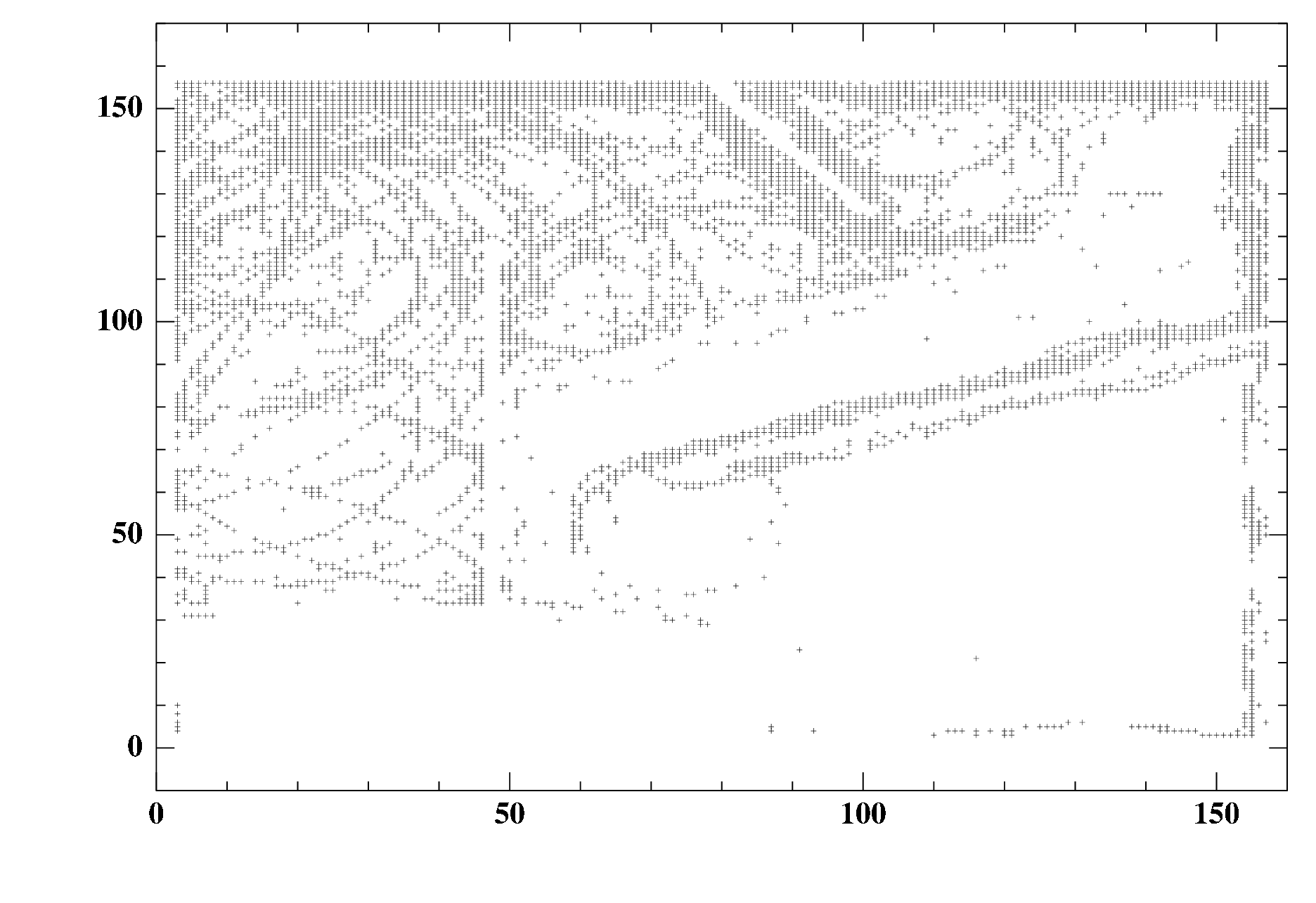}\\
\includegraphics[width=1.1in]{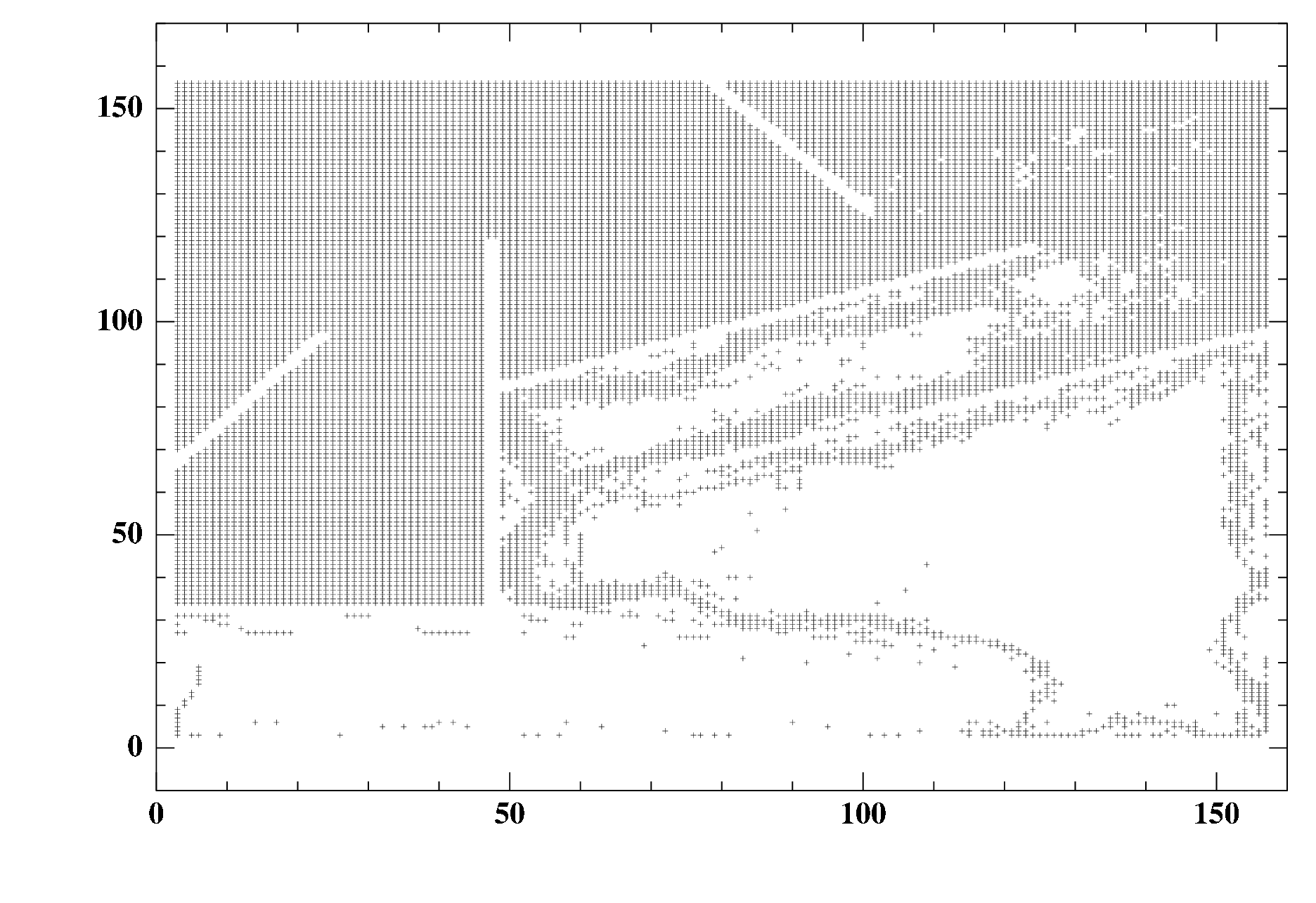} &
\includegraphics[width=1.1in]{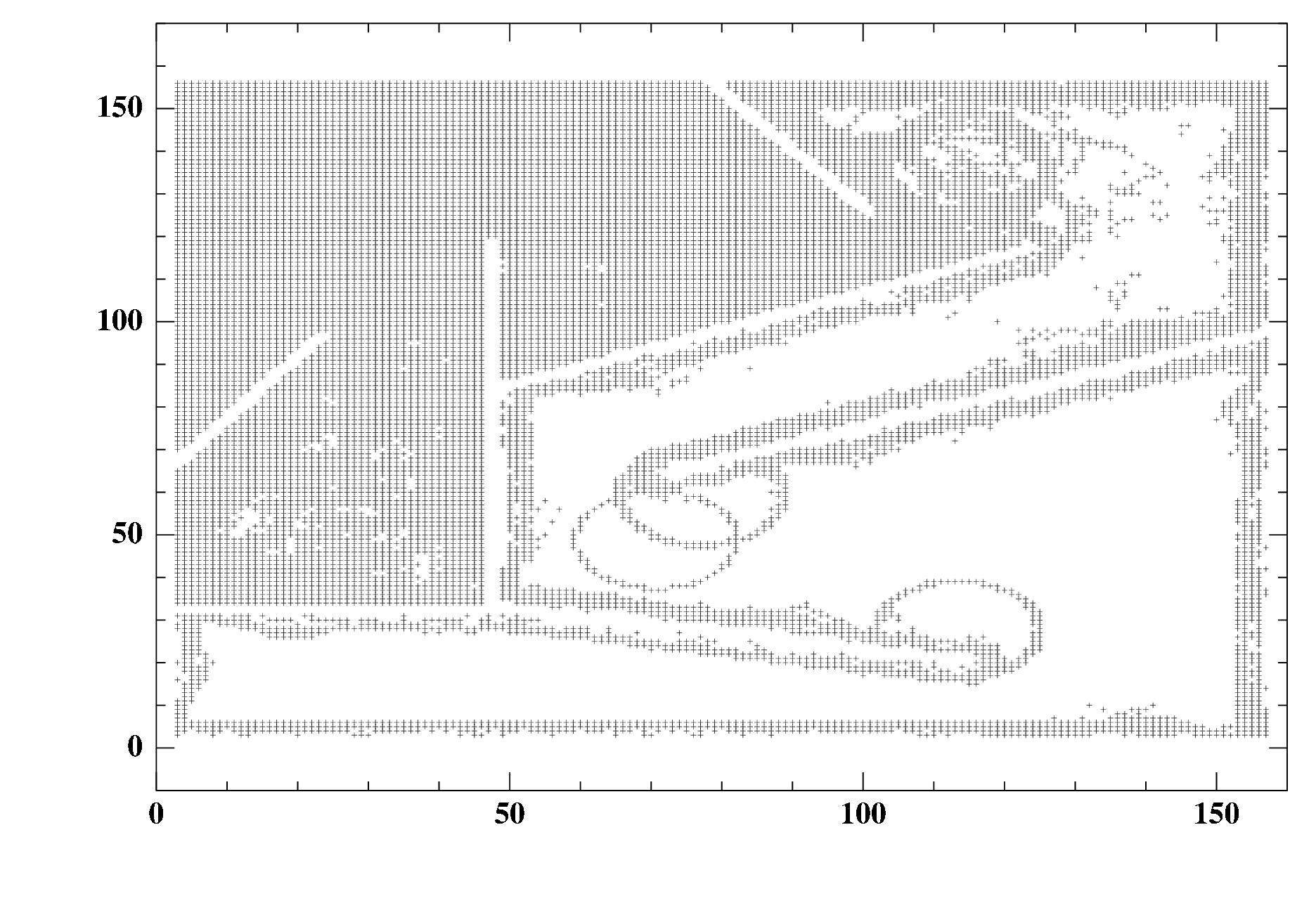} &
\includegraphics[width=1.1in]{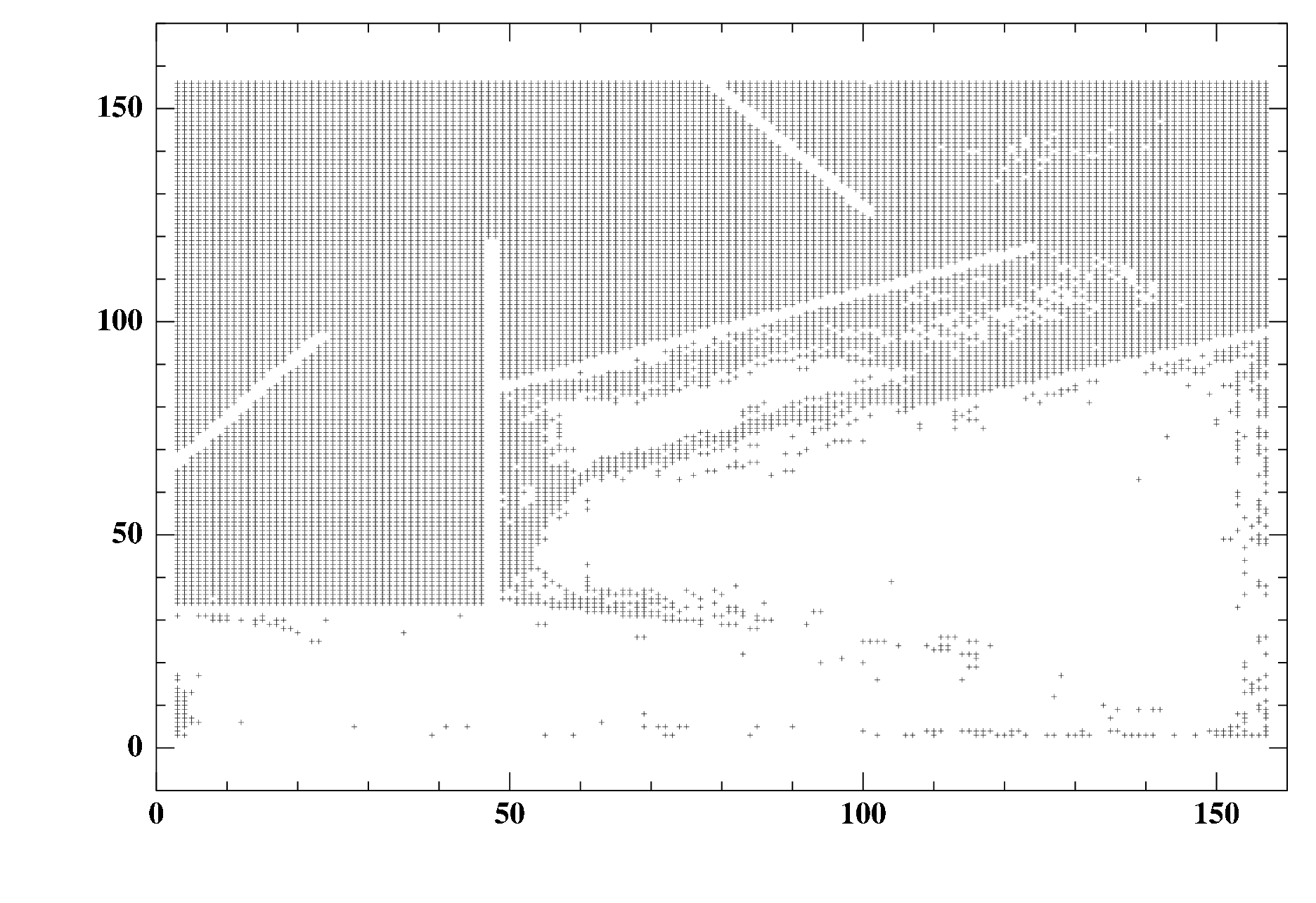} &
\includegraphics[width=1.1in]{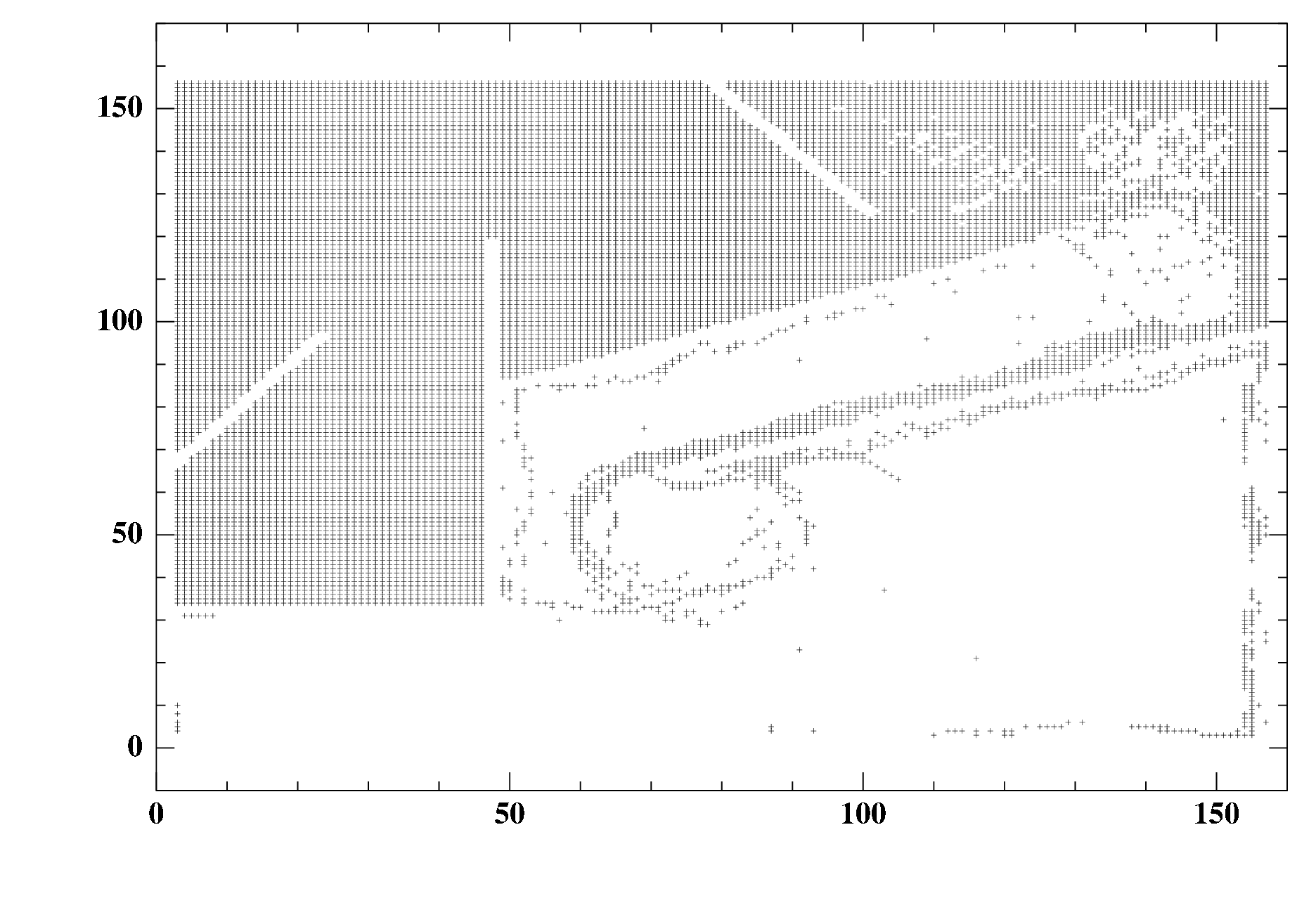} \\
Curiosity & Discovery & \NF & \LS \\
\end{tabular}}
\caption{Hard  Arena: points visited 10 times or more by the 100 best individuals (top) or the whole population (bottom) over the 11 runs,}
\label{fig:aren}
\end{figure}

The average patrolling abilities for both arenas and the 4 algorithms are reported in Table \ref{tab:patrol}: for the 100 best individuals that were evolved during the 11 runs (top), and all the individuals that appeared during those runs (bottom), Table \ref{tab:patrol} gives the percentage of the arena that has been visited at least 2, 5 or 10 times. Fig. \ref{fig:aren} displays a typical case: the points that have been visited at least 10 times in the Hard arena setting (the Medium arena shows similar trends). 

When considering the 100 best individuals that appeared during the 11 runs, Novelty is clearly outperforming the other approaches in terms of patrolling  the Medium arena, while Curiosity (and, to a lesser extend, Displacement) catch up in the co-called Hard arena. Indeed, and almost paradoxically, the Hard arena offers in fact a high sensor diversity, exhibiting zones that look rather different, while the Medium arena somehow repeats the same motif several times, generating less diverse sensor input combinations. Furthermore, looking at the plots of Fig. \ref{fig:aren}, the Novelty runs tends to also explore the interior regions of the maze, while the Curiosity runs stay along the walls: the empty zones always generates the same sensor values, and hence cannot contribute to increase the entropy. However, as can be seen on Fig. \ref{fig:aren}, and is confirmed by looking at the maximum distance from the starting position reached by the individuals of different generations (results not shown here), far chambers from the starting points are more densely visited by the Curiosity runs than by the Novelty ones (the first wall from bottom up is more clearly marked in the Curiosity plot).\\
Finally, the Discovery fitness performs poorly when considering the 100 best individuals. Indeed, those individuals essentially correspond to the 100 last individuals, visiting sensori-motor states which have not been visited by the ancestors, hence basically exploring only the corners of the arena. 

When considering all individuals ever produced by evolution during the 11 runs (bottoms half of Table \ref{tab:patrol} and bottom row of Fig. \ref{fig:aren}), the picture changes dramatically. The best performing fitness now is Curiosity, and again this is even clearer on the Hard arena. The Displacement fitness now also reaches better performances than Novelty, again more clearly on the Hard arena. Finally, the performance of Discovery is almost as good as the other ones, and the plot of the 10-times visits is even denser toward the far end of the arena with respect to the starting point, i.e., close to the bottom wall: because of the inherited information from parents to offspring, Discovery should indeed only be assessed by looking at complete lineages. \\
The success of Curiosity compared to Novelty is somewhat unexpected, as Novelty actually relies on a significant amount of prior knowledge, requiring for instance the robot to always know its position, and, from the archive, where all other robots ended up their trajectories. On the opposite, the Curiosity fitness is built up from scratch by each robot -- or by the lineage of robot in the case of the Discovery fitness.

\section{Discussion and Perspectives}
The paper pioneers the use of statistical unsupervised learning to define self-driven
fitness functions for on-board ER. Information Theory is then used to push the robot toward unknown parts of the sensori-motor space, hopefully leading to interesting behaviors in the physical space. 

The priority here is to enable the robot to merely discover
its sensori-motor space, as opposed to maximize the predictive information in the sensori-motor loop
(as in \cite{Zahedi2010} and references therein). The rationale for this priority is twofold. Firstly, the presented approach is extremely frugal computationally speaking, compared to e.g. \cite{Zahedi2010}, as the target application here concerns swarm robotics. Secondly, recent trends in Machine Learning suggest that changing the representation of the problem domain, as done through unsupervised learning, can dramatically facilitate further supervised learning tasks.

Compared to the standard 
Evolutionary Robotics framework pioneered by Floreano and Mondada, the robot is rewarded here for what it gets (a rich sensori-motor experience) and not for what it does (going fast and circling infrequently).  
As mentioned earlier on, such a fitness function is only efficient in ``interesting environments''; 
under-stimulation results in a evolutionary bootstrap problem. Interestingly, over-stimulation is also 
detrimental to the efficiency of the curiosity and discovery instincts.

Compared to the Novelty approach by Lehman and Stanley \cite{Lehman}, no external information is needed here to assess the novelty of a behaviour. Furthermore, almost independently of the context, and the goal (though Novelty search can be totally goal-less, it can also be constrained toward a loosely defined goal), the computational cost of the proposed approaches remains tractable, as only sensori-motor states need to be stored.\\
Compared to Embodied Statistical Learning and Intrinsic Motivation \cite{Oudeyer2}, beside being computationally light learning algorithms amenable to on-board evolution, the proposed approaches rely on the discovery of sensori-motor states, amenable
to the direct inspection of the designer. Typically, relating visited sms to some instants of the 
trajectory would lead to interpreting them, thus allowing the designer
to enforce some preferences, e.g., a safety policy in critical situations. 

Further study will of course concern experiments in richer behavioral spaces, and also the collective and dynamic adjustment of the 
clustering granularity $\epsilon$, depending on the current context.  Another perspective is related to
coupling self-driven fitnesses with interactive optimization, asking the designer's preferences
among the available behaviours.


\end{document}